%% file: 0_main.tex
\documentclass[10pt,twocolumn,letterpaper]{article}

\usepackage{wacv}
\usepackage{times}
\usepackage{epsfig}
\usepackage{graphicx}
\usepackage{amsmath}
\usepackage{amssymb}
\usepackage{floatrow}
\usepackage{subcaption}
\usepackage{algorithmicx,algorithm}
\usepackage[noend]{algpseudocode}
\usepackage[accsupp]{axessibility}
\newenvironment{densitemize}
{\begin{list}               
    {$\bullet$ \hfill}{
        \setlength{\leftmargin}{\parindent}
        \setlength{\parsep}{0.04\baselineskip}
        \setlength{\itemsep}{0.5\parsep}
        \setlength{\labelwidth}{\leftmargin}
        \setlength{\labelsep}{0em}}
    }
{\end{list}}

\def\wacvPaperID{1068} 

\wacvfinalcopy 

\ifwacvfinal
\usepackage[breaklinks=true,bookmarks=false]{hyperref}
\else
\usepackage[pagebackref=true,breaklinks=true,colorlinks,bookmarks=false]{hyperref}
\fi

\ifwacvfinal
\fi

\begin{document}

\title{Online Continual Learning Via Candidates Voting}

\author{Jiangpeng He\\
{\tt\small he416@purdue.edu}
\and

Fengqing Zhu\\
{\tt\small zhu0@purdue.edu}
\and
{School of Electrical and Computer Engineering, Purdue University, West Lafayette, Indiana USA}
}

\maketitle
\thispagestyle{empty}

\begin{abstract}

Continual learning in online scenario aims to learn a sequence of new tasks from data stream using each data only once for training, which is more realistic than in offline mode assuming data from new task are all available. However, this problem is still under-explored for the challenging class-incremental setting in which the model classifies all classes seen so far during inference. Particularly, performance struggles with increased number of tasks or additional classes to learn for each task. In addition, most existing methods require storing original data as exemplars for knowledge replay, which may not be feasible for certain applications with limited memory budget or privacy concerns. In this work, we introduce an effective and memory-efficient method for online continual learning under class-incremental setting through candidates selection from each learned task together with prior incorporation using stored feature embeddings instead of original data as exemplars. Our proposed method implemented for image classification task achieves the best results under different benchmark datasets for online continual learning including CIFAR-10, CIFAR-100 and CORE-50 while requiring much less memory resource compared with existing works.

\end{abstract}


\input{1_introduction}

\input{2_related_work}



\input{4_method_our}

\input{5_experimental_results}

\input{6_conclusion}

{\small
\bibliographystyle{ieee_fullname}
\bibliography{egbib}
}
\newpage
\input{7_supplementary}

\end{document}

%% file: 1_introduction.tex
\section{Introduction}
\label{introduction}
Continual learning, a promising future learning strategy, is able to learn from a sequence of tasks incrementally using less computation and memory resource compared with retraining from scratch whenever observing a new task. However, it suffers from catastrophic forgetting~\cite{CF}, in which the model quickly forgets already learned knowledge due to the unavailability of old data. Existing methods address this problem under different scenarios including (1) \textit{task-incremental} vs. \textit{class-incremental} depending on whether task index is available and (2) \textit{offline} vs. \textit{online} depending on how many passes are allowed to use each new data. In general, \textit{online class-incremental} methods use each data once to update the model and employs a single-head classifier~\cite{SIT} to test on all classes encountered so far during inference. This setting is more closer to real life learning environment where new classes come in as data streams with limited adaptation time and storage capacity allowed for processing~\cite{mai2021online}. Unfortunately, class-incremental learning in online scenario is not well-studied compared with offline setting. In addition, existing online methods~\cite{A-GEM,shim2020online_ASER,prabhu2020gdumb_online,ILIO,aljundi_mir} all require original data from each learned task as exemplars, which restricts their deployment for certain applications (\textit{e.g.}, healthcare and medial research) with memory constraints or privacy concerns. Therefore, an effective online continual learning method is needed to address the above challenges for real world deployment and to improve the performance of online methods. 

\begin{figure}[t]
\begin{center}
  \includegraphics[width=1.\linewidth]{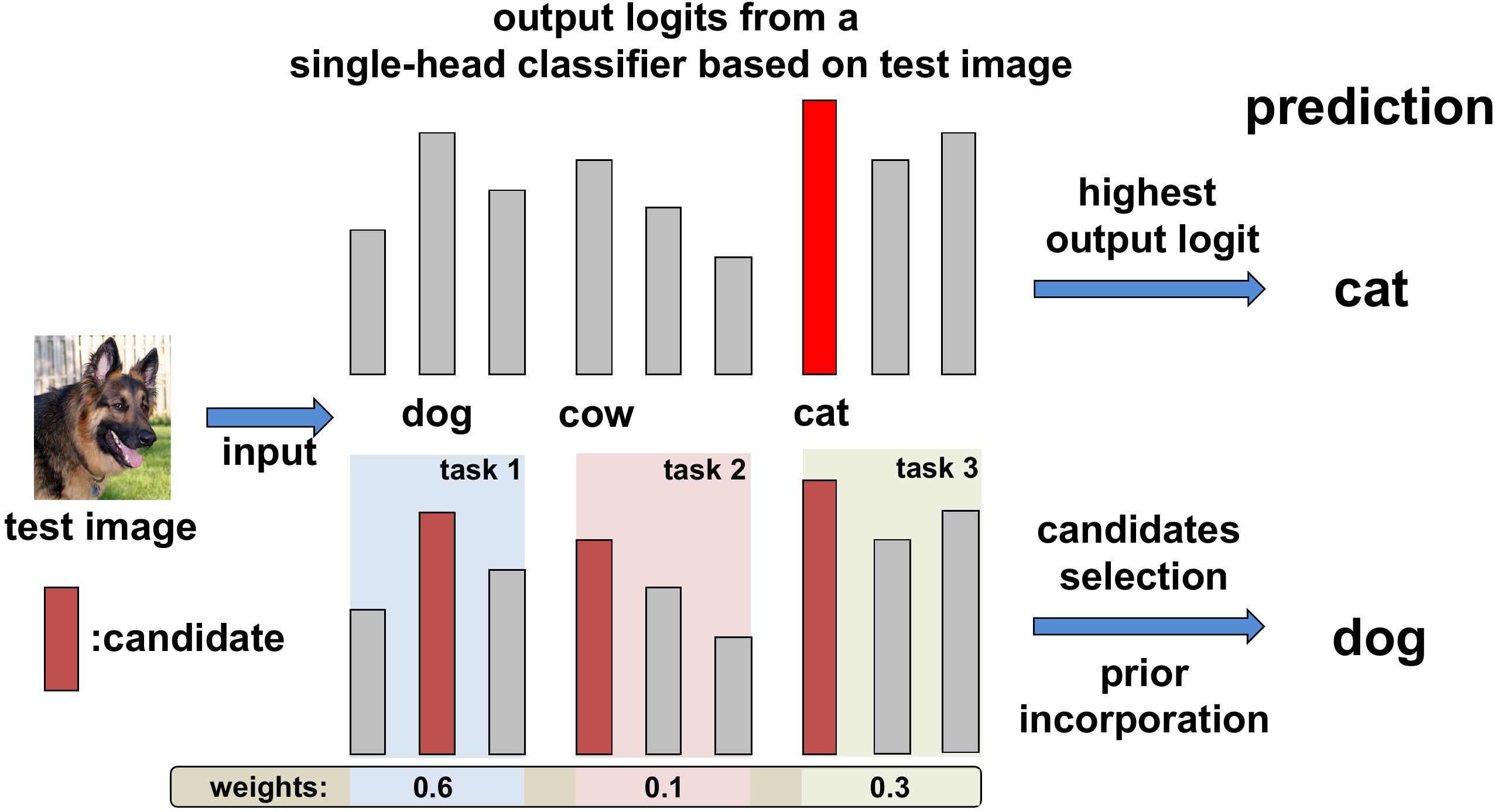}
  \vspace{-0.9cm}
  \caption{\textbf{Illustration of the difference between our proposed method and other methods to make prediction based on output of a single-head classifier.} With single-head classifier, the output is associated with the largest value of the output logits. In contrast, our method makes prediction by first selecting candidates from each learned task and then incorporating the corresponding weights. }
  \label{fig:intro_diff}
\end{center}
\end{figure}



For class-incremental methods using a single-head classifier, the prediction result is always associated with the largest value of output logits. However, during continual learning, the output logits become biased towards new task due to the unavailability of old task data~\cite{BIC}, \textit{i.e.}, the output logits of new task are much larger than those of old tasks. This results in the corresponding biased prediction on new tasks, which is a significant contributing factor for catastrophic forgetting. Our method is motivated by the observation that the model is still able to maintain its discriminability for classes within each task~\cite{mainatining} despite the bias issue, \textit{i.e.}, the correct class label can be drawn from the candidate prediction given by each learned task during inference. Therefore, our method aims to treat the class label associated with the largest output logit for each learned task as a candidate and the final prediction is based on the weighted votes of all selected candidates. Figure~\ref{fig:intro_diff} illustrates the main difference between our method and others to make prediction based on the output of a single-head classifier.

To achieve this goal, there are two associated questions we need to address: (1) How to obtain the largest logits as candidates from the output of each learned task using a single-head classifier without knowing the task index? (2) How to generate the weight for each selected candidate to determine the final prediction? In this work, we address both problems by leveraging exemplar set~\cite{ICARL}, where a small number of old task data is stored for replay during continual learning. However, different from existing methods~\cite{A-GEM,shim2020online_ASER,prabhu2020gdumb_online,ILIO,aljundi_mir} which use original data as exemplar, we apply a feature extractor and store only feature embeddings, which is more memory-efficient and privacy-preserving. We argue that the task index can be stored together with selected exemplars while learning each new task. Therefore, during inference phase, we can directly obtain the output logits for each learned task from the single-head classifier based on stored task index in the exemplar set and extract the largest output logits. We refer to this as the \textbf{candidates selection} process. In addition, we design a probabilistic neural networks~\cite{PNN} leveraging all stored feature embeddings to generate the probability distribution of learned task that the input test data belongs to, and use it as the weights to decide the final prediction. We denote this step as \textbf{prior incorporation}.
The main contributions are summarized as follows.
\begin{densitemize}
    \item We propose a novel and efficient framework for online continual learning through candidates selection and prior incorporation without requiring original data to reduce the memory burden and address privacy issue for real world applications.

    \item An online sampler is designed to select exemplars from sequentially available data stream through dynamic mean update criteria and we further study exemplar augmentation in feature space to achieve improved performance
    
    \item We conduct extensive experiments on benchmark datasets including CIFAR-10~\cite{CIFAR}, CIFAR-100~\cite{CIFAR} and CORE-50~\cite{core50} and show significant improvements compared with existing online methods while requiring the least storage. 
    
   \item We further show that our online method outperforms state-of-the-art offline continual learning approaches on CIFAR-100~\cite{CIFAR} dataset, at the same time it alleviates the weight bias problem and reduces the memory storage consumption compared with existing works.
\end{densitemize}


%% file: 2_related_work.tex
\begin{figure*}[htbp]
\begin{center}
  \includegraphics[width=.9\linewidth]{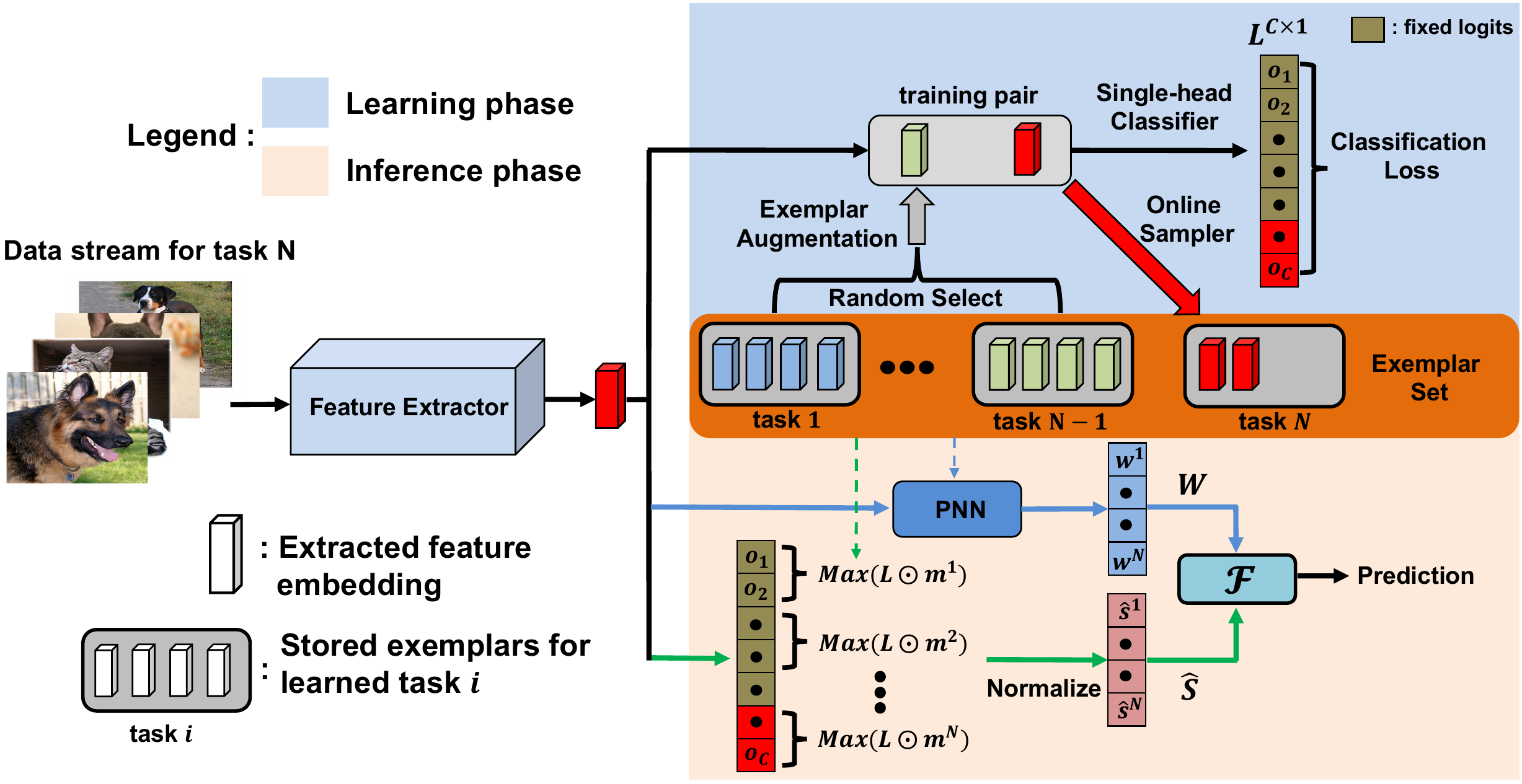}
  \caption{\textbf{Overview of our proposed online continual learning method to learn a new task $N$.} The upper half shows the learning phase where we pair the extracted feature of new data with an exemplar to train the single-head classifier. $L$ denotes the output logits for all classes $C$ seen so far. The parameters for each learned task in the classifier are fixed to maximally maintain its discriminability and an online sampler is designed to select exemplars for current task $N$. The lower half shows the inference phase where the candidates selection and prior incorporation are denoted by green and blue arrows, respectively. The output logits for each learned task is obtained using element-wise product on classifier output $L$ and binary mask $\{m^{i}, i = 1,2,...N\}$ generated from exemplar set and we treat the highest logits for each task as candidates. A probabilistic neural network (PNN) is designed using all stored exemplars to provide the prior information of which task index the input data belongs to during inference, which can be regarded as weights for selected candidates to obtain the final prediction using our proposed function $\mathcal{F}$. (Best viewed in color)}
  \label{fig:overall}
  \vspace{-0.4cm}
\end{center}
\end{figure*}

\section{Related Work}
\label{related work}
Continual learning is studied under different learning scenarios. In general, it can be divided into (1) class-incremental (2) task-incremental and (3) domain-incremental as discussed in~\cite{hsu2018re}. Instead of using a single-head classifier~\cite{SIT} for all classes seen so far in class-incremental setting, methods for task-incremental problem apply a multi-head classifier~\cite{abati2020conditional_taskaware} for each independent task and domain-incremental methods aim to learn the label shift rather than new classes. In addition, depending on whether each data is used more than once to update model, it can be categorized as (1) online learning that use each data once and (2) offline learning with no epoch restriction. In this work, we study the continual learning under online and class-incremental setting, where the model observes each data once and perform classification within all seen classes during inference phase. In this section, we review existing continual learning works related to our method in two categories including (1) Regularization-based and (2) Replay-based methods. 

\textit{Regularization-based} methods restrict the impact of learning new tasks on the parameters that are important for learned tasks. Representative methods include freezing part of layers~\cite{jung2016_lessforgetting, kirkpatrick2017overcoming} and using distillation loss or its variants~\cite{LWF,ILIO,EEIL,ICARL,R_and_D,rebalancing, inthewild, he2021online_food}. However, they also limit the model's ability to learn new task and can even harm the performance if the teacher model used by distillation~\cite{KD} is not learned on large balanced data~\cite{dualmemory}. Our method applies a fixed backbone model that is pre-trained on large scale datasets to extract feature embeddings of new data as input and uses cross-entropy to learn a discriminative classifier for each new task. Therefore, even though we freeze the parameters for learned tasks in the classifier, it has minimum impact on extracted features to learn new task. Recent studies~\cite{BIC,mainatining} also found that the bias of model weights towards new classes is one of the reasons for catastrophic forgetting. Therefore, Wu \textit{et al}.~\cite{BIC} proposed to correct the weights by applying an additional linear model. Then Weight Aligning is proposed in~\cite{mainatining} to directly correct the biased weights in the FC layer without requiring additional parameters. However, none of these methods are designed for online scenario where each data is only allowed to use once for training. In this work we propose to tackle this problem from a novel perspective by selecting candidates for each learned task and then use the weighted score for final prediction, which effectively addresses catastrophic forgetting in online case. 

\textit{Replay-based} methods are shown to be effective for maintaining learned knowledge by either using the original data as exemplars~\cite{ICARL,prabhu2020gdumb_online,GEM,EEIL,liu2020mnemonics,pomponi2020efficient,aljundi_mir,NEURIPS2019_e562cd9c_GSS,A-GEM, shim2020online_ASER, chaudhry2019_TINYER, chaudhry2020using_HAL} or synthetic data and statistics~\cite{shin2017continual, venkatesan2017strategy,cgan, fearnet}. However, using original data may not be feasible for certain applications due to privacy concerns and also it may require large storage depending on the size of input data. In addition, using synthetic data or data statistic require training a generative model~\cite{GAN} during learning phase, which is not feasible in online scenario. Therefore, we propose to use feature embeddings as exemplars for rehearsal to mitigate forgetting in online case. Besides, we also utilize the stored feature to (1) generate binary masks for each learned task to select candidates and (2) provide prior information as weights to obtain final prediction. We argue that both information are valuable to explore, particularly under the online continual learning context when available resource is limited. 

Among these methods, only a few are studied for online mode~\cite{GEM, prabhu2020gdumb_online,aljundi_mir,NEURIPS2019_e562cd9c_GSS, shim2020online_ASER, A-GEM, chaudhry2019_TINYER, chaudhry2020using_HAL} with even less work under class-incremental setting~\cite{prabhu2020gdumb_online, NEURIPS2019_e562cd9c_GSS, aljundi_mir, shim2020online_ASER}, which is more challenging but also worth investigating as it closely relates to applications in real world scenario.


%% file: 4_method_our.tex
\section{Our Method}
\label{method:our}
The overview of our method is illustrated in Figure~\ref{fig:overall}, including a learning phase to learn new task from a data stream and an inference phase to test for all tasks seen so far. Our method applies a fixed backbone network to extract feature embedding as input, which is more discriminative, memory-efficient and also privacy-preserving compared with using original data. We freeze the parameters in the classifier after learning each new task to maximally maintain its discriminability. We emphasize that our method still uses a single-head classifier but restricts the update of parameters corresponding to all learned tasks. 

\subsection{Learning Phase}
\label{method: learning phase}
The upper half of Figure~\ref{fig:overall} shows the learning phase in online scenario where we train the classifier by pairing each extracted feature embedding of the new data with one exemplar randomly selected from exemplar set into the training batch. Cross-entropy is used as the classification loss to update the model, which generates a more discriminative classifier as no regularization term on learned tasks is used. It also does not require additional memory to store the output logits compared with using knowledge distillation loss~\cite{KD}. 

\begin{algorithm}[t]
\caption{Online Sampler}
\begin{flushleft}
    \hspace*{0.02in} {\bf Input:}
    Data stream for task N: $\{(\textbf{x}_1,y_1)^N,(\textbf{x}_2,y_2)^N, ...\}$\\
    \hspace*{0.02in} {\bf Require:} 
    Backbone feature extractor $\mathcal{F}$\\
    \hspace*{0.02in} {\bf Output:} 
    Updated exemplar set: $E^{N-1} \rightarrow E^{N}$
\end{flushleft}
\vspace{-0.35cm}
\begin{algorithmic}[1]
\For{i = 1, 2, ... }
\State $v_i \leftarrow \mathcal{F}(\textbf{x}_i)$ \Comment{\small{Extract feature embedding}} 
\State $f_m^{(y_i)} \leftarrow \frac{n_{y_i}}{n_{y_i}+1}f_m^{(y_i)} + \frac{1}{n_{y_i}+1}v_i$ \Comment{\small{online mean update}}
\State $n_{y_i} \leftarrow n_{y_i} + 1$ \Comment{total number of seen data}
\If {$|E^N(y_i)| < q$} \Comment{\small{exemplars for class $y_i$ not full}}
\State $E^N(y_i) \leftarrow E^N(y_i) \cup (\textbf{v}_i,y_i)^N$
\Else
\State $ I_{max} \leftarrow \textit{argmax}(||v_j -f_m^{(y_i)}||^2, j \in i \cup E^N(y_i))$
\EndIf
\If{$I_{max} \neq i$}
\State Remove $(\textbf{v}_{I_{max}}, y_i)^N$ from $E^N(y_i)$
\State $E^N(y_i) \leftarrow E^N(y_i) \cup (\textbf{v}_i,y_i)^N$
\Else
\State \textbf{Continue}
\EndIf
\EndFor
\end{algorithmic}
\label{alg:online sampler}
\end{algorithm} 

\textbf{Online sampler: }There are two necessary conditions we need to satisfy when designing the online sampler for our method: (1) it should be able to select exemplars from sequentially available data in online scenario, (2) the selected exemplars should near the class mean as we will leverage stored features to provide prior information using distance-based metric during inference phase, which is described later in Section~\ref{method:inference}. However, none of the existing exemplar selection algorithms satisfy both conditions. In addition, although Herding~\cite{HERDING} is widely applied to select exemplars based on class mean, it only works in offline scenario assuming the data from new task is all available. Therefore, we propose to use an online dynamic class mean update criteria~\cite{guerriero2018deepncm} for exemplar selection, which does not require knowing the total number of data beforehand as shown in Equation~\ref{eq:onlinemean}. 
\begin{equation}
\label{eq:onlinemean}
\textbf{v}_{mean} = \frac{n}{n+1} \textbf{v}_{mean} + \frac{1}{n+1} \textbf{v}_{n}
\end{equation}
where $n$ refers to the number of data seen so far in this class and $\textbf{v}_{n}$ denotes a new observation. Algorithm~\ref{alg:online sampler} illustrates the exemplar selection process for a new task $N$, where $q  = \frac{Q}{|class|}$ denotes the number of allowable exemplars per class given total capacity $Q$ and $f_m^{(y_i)}$ is the mean vector for total $n_{y_i}$ data seen so far for class label $y_i$. The exemplar set can be expressed as $E=\{(\textbf{v}_1,y_1)^1,(\textbf{v}_2,y_2)^1,..., (\textbf{v}_1,y_1)^N,(\textbf{v}_2,y_2)^N,...\}$, where $(\textbf{v}_j,y_j)^k$ denotes the $j$-th stored exemplar for the $k$-th learned task and $k\in \{1,2,...,N\}$. Each stored exemplar contains extracted feature $\textbf{v}$, class label $y$ and task index $k$. 

\textbf{Exemplar augmentation in feature space: }Although exemplars help to remember learned tasks by knowledge reply during continual learning, the model performance greatly depends on the size of the exemplar set, \textit{i.e.,} the larger the better, which is challenging given a limited memory budget particularly in online scenario. Therefore, we also study the exemplar augmentation techniques in this work to help improve the performance without requiring additional storage. Since we store feature embedding as exemplar, common data augmentation methods that are typically applied to image data such as rotation, flip and random crop cannot be used directly in feature space. Therefore, we adopt random perturbation for feature augmentation~\cite{FSDA}.

\textit{Random perturbation: }We generate pseudo feature exemplar by adding a random vector $P$ drawn from a Gaussian distribution with zero mean and per-element standard deviation $\sigma$ as shown in Equation~\ref{eq:random_per}
\begin{equation}
\label{eq:random_per}
\Tilde{\textbf{v}}_i = \textbf{v}_i + \alpha_r P, \quad P\sim N(0,\sigma_i)
\end{equation}
where $\textbf{v}_i$ refers to the stored feature in exemplar set, and $\Tilde{\textbf{v}}_i$ denotes the augmented feature. $\alpha_r$ is a constant which controls the scale of noise, and is set to $\alpha_r = 1$ in our implementation. We emphasize that we do not need to store augmented feature in exemplar set and the exemplar augmentation is randomly implemented when pairing the extracted feature of new data. 

\begin{figure*}[t]
\centering
\begin{minipage}[t]{0.22\linewidth}
    \centering
    \includegraphics[width=4.cm,height=3.5cm]{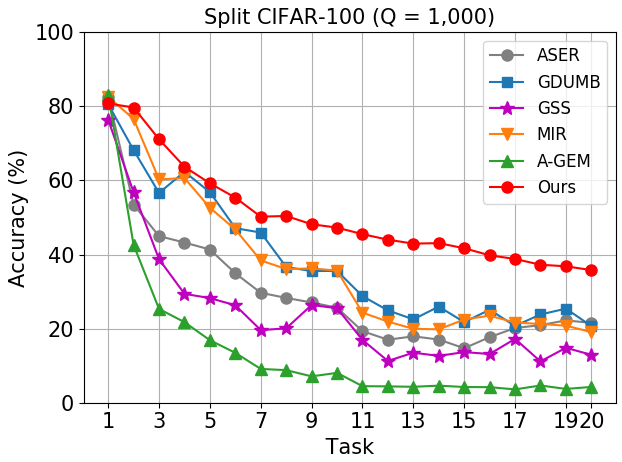}
    \parbox{15.5cm}{\small \hspace{2.cm}(a)}
\end{minipage}
\hspace{2ex}
\begin{minipage}[t]{0.22\linewidth}
    \centering
    \includegraphics[width=4.cm,height=3.5cm]{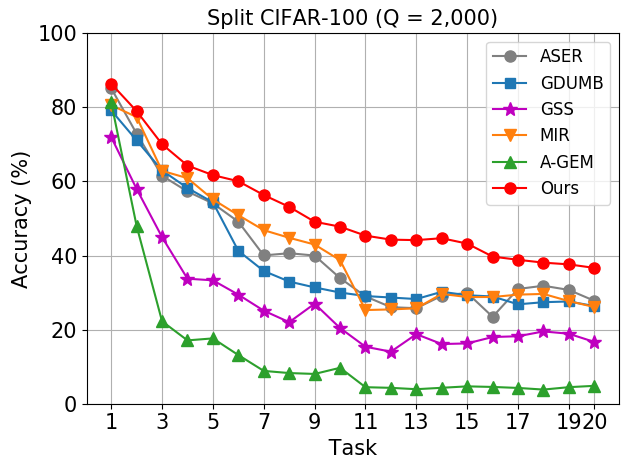}
    \parbox{15.5cm}{\small \hspace{2.cm}(b)}
\end{minipage}
\hspace{2ex}
\begin{minipage}[t]{0.22\linewidth}
    \centering
    \includegraphics[width=4.cm,height=3.5cm]{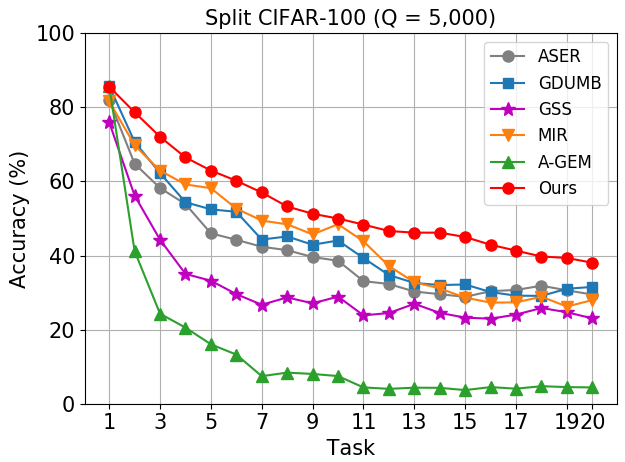}
    \parbox{15.5cm}{\small \hspace{2.cm}(c)}
\end{minipage}
\hspace{2ex}
\begin{minipage}[t]{0.22\linewidth}
    \centering
    \includegraphics[width=4.cm,height=3.5cm]{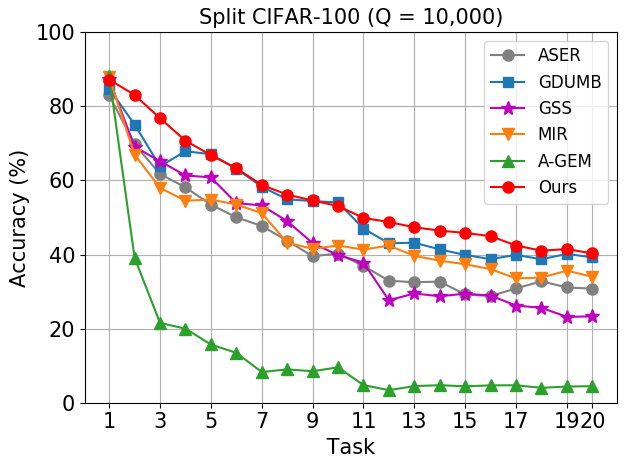}
    \parbox{15.5cm}{\small \hspace{2.cm}(d)}
\end{minipage}
\begin{center}
\vspace{-.9cm}
\caption{\textbf{Results on Split CIFAR-100} by comparing with existing online methods with different exemplar size $Q$. The accuracy is measured after learning of each task on all tasks seen so far. (Best viewed in color)}
\label{fig: online result on cifar}
\end{center}
\end{figure*}


\begin{table*}[t!]
    \vspace{-0.1cm}
    \centering
    \scalebox{.85}{
    \begin{tabular}{ccccccccc|cccccccc}
        \hline
        Datasets & \multicolumn{8}{c}{\textbf{Split CIFAR-10}}& \multicolumn{8}{c}{\textbf{CORE-50}}  \\
        \hline
        Size of exemplar set & \multicolumn{2}{c}{$Q = 1,000$} & \multicolumn{2}{c}{$Q = 2,000$} & \multicolumn{2}{c}{$Q = 5,000$} & \multicolumn{2}{c}{$Q = 10,000$}& \multicolumn{2}{c}{$Q = 1,000$} & \multicolumn{2}{c}{$Q = 2,000$} & \multicolumn{2}{c}{$Q = 5,000$} & \multicolumn{2}{c}{$Q = 10,000$}\\
        \hline
        Accuracy(\%) & Avg & Last & Avg & Last & Avg & Last & Avg & Last & Avg & Last & Avg & Last & Avg & Last & Avg & Last \\
        \hline
        \hline
        A-GEM~\cite{A-GEM}   & 43.0 & 17.5 & 59.1 & 38.3 & 74.0 & 59.0 & 74.7 & 62.5 & 20.7 & 8.4 & 21.9 & 10.3 & 22.9 & 11.5 & 24.6 & 12.0 \\
        MIR~\cite{aljundi_mir}  & 67.3 & 52.2 & 80.2 & 66.2 & 83.4 & 74.8 & 86.0 & 78.4 & 33.9 & 21.1 & 37.1 & 24.5 & 38.1 & 27.7 & 41.1 & 31.8 \\
        GSS~\cite{NEURIPS2019_e562cd9c_GSS}  & 70.3 & 56.7 & 73.6 & 56.3 & 79.3 & 64.4 & 79.7 & 67.1 & 27.8 & 17.8 & 31.0 & 18.9 & 31.8 & 21.1 & 33.6 & 22.6 \\
        ASER~\cite{shim2020online_ASER} & 63.4 & 46.4 & 78.2 & 59.3 & 83.3 & 73.1 & 86.5 & 79.3 & 24.3 & 12.2 & 30.8 & 17.4 & 32.5 & 18.5 & 34.1 & 21.8\\
        GDUMB~\cite{prabhu2020gdumb_online} & 73.8 & 57.7 & 83.8 & 72.4 & 85.3 & 75.9 & 87.7 & 82.3 & 41.2 & 23.6 & 48.4 & 32.7 & 54.3 & 41.6 & 56.1 & 45.5\\
        \hline
        Ours &  \textbf{76.0} & \textbf{62.9} & \textbf{84.9} & \textbf{74.1} & \textbf{86.1} & \textbf{77.0} & \textbf{88.3} & \textbf{82.7} & \textbf{45.1} & \textbf{26.5} & \textbf{50.7} & \textbf{34.5} & \textbf{56.3} & \textbf{43.1} & \textbf{57.5} & \textbf{46.2} \\
        \hline
    \end{tabular}
    }
    \vspace{-0.2cm}
    \caption{\textbf{Average accuracy and Last step accuracy on Split CIFAR-10 and CORE-50}. Best results marked in bold.  }
    \label{tab:online_cifar_core}
\end{table*}


\vspace{-0.1cm}
\subsection{Inference Phase}
\vspace{-0.1cm}
\label{method:inference}
The lower half of Figure~\ref{fig:overall} shows inference phase, which comprises of two key components: candidates selection and prior incorporation. The stored exemplars along with their task indexes are used to generate binary mask to obtain the corresponding output logits for each learned task during inference. We extract the highest output as candidates and a variant of probabilistic neural network (PNN)~\cite{PNN} using all stored exemplars is designed to provide prior information as weights for selected candidates to vote for final prediction, which will be described in detail below. 

\textbf{Candidates selection: }We denote  $L = \{o^1, o^2, ..., o^C \}$ as the output logits from the single-head classifier where $C$ refers to the total number of seen classes belonging to $N$ learned tasks so far. During inference phase, the exemplar set generates a binary mask $m^k \in \{0, 1\}^{C}$ for task $k$ by assigning the $i$-th entry $m^k_i$ as $1$ if class label $i$ belongs to task $k$ and as $0$ if not, so we have $\sum^C_{i=1}m^k_i = C^k$, where $C^k$ is the number of classes belonging to task $k$. Thus, the candidate output logit from each learned task is selected by
\begin{equation}
\label{eq:candidate selection}
s^k = \textit{Max} \{L \odot m^k \}, \quad k = 1,2,...,N
\end{equation}
where $\odot$ refers to element-wise product. We then perform normalization step for the extracted candidate logits by using the corresponding norm of weight vectors in classifier. Specifically, for each selected candidate $s^k$, let $W^k \in \mathcal{R}^{d_m \times 1}$ and $|W^k|$ denotes the weight vector in classifier and its norm respectively where $d_m$ is the input dimension. Then we normalize each candidate with $$\hat{s}^k = \frac{1}{|W^k|} \frac{s^k - \textit{Min}\{s^1, ...s^N\} }{\epsilon_n+\sum_{j=1}^N (s^j - \textit{Min}\{s^1, ...s^N\})}$$ where $\epsilon_n$ is for regularization and larger $\hat{s}$ can reflect higher probability as prediction.
Finally, the normalized selected candidates for $N$ learned tasks can be expressed as $\hat{S} = \{\hat{s}^1, \hat{s}^2, ..., \hat{s}^N\}$ with corresponding extracted candidate class labels $Y = \{y^1, y^2, ..., y^N\}$. 

\textbf{Prior incorporation: }We apply PNN to generate prior probability distribution of which learned task index the test data belongs to. PNN computes class conditional probabilities using all stored features in the exemplar set. Specifically, it calculates the probability that an input feature vector $\textbf{f}$ belongs to task $k$ as formulated in Equation~\ref{eq:pnn} below.
\begin{equation}
    \label{eq:pnn}
    \begin{split}
             & P(k|\textbf{f}) = \frac{\alpha^k}{\sum_{i=1}^N\alpha^i} \\ 
            & \alpha^k = (\epsilon_r + \textit{Min}_j||\textbf{f} - \textbf{v}_j^k||_2) )^{-1}
    \end{split}
\end{equation}
where $\epsilon_r > 0$ is used for regularization and $\textbf{v}_j^k$ denotes the $j$-th stored feature in exemplar set for learned task $k$.

The output of PNN is a $N$ dimension prior vector $W = (w^1,w^2,...,w^N) $ and we use it as the weights to combine with the normalized candidates $\hat{S}$ to get final predicted class label $\hat{y}$ using Equation~\ref{eq:combine}.
\begin{equation}
    \label{eq:combine}
    \hat{y} = \mathop{\textit{argmax}}_{y^i \in Y}(\hat{s}^i + e^{(\gamma-1)}\times w^i)
\end{equation}
where $\gamma = \frac{\textit{Max}(W)-\textit{Min}(W)}{\beta}$ is a dynamic hyper-parameter used for incorporation determined by calculating difference between maximum and minimum value in prior vector. $\beta \in (0,1)$ is a normalization constant. In this work, we show the effectiveness of our method by using a fixed $\beta = 0.5$ for all experiments. 

%% file: 5_experimental_results.tex
\begin{figure*}[t]
\centering
\begin{minipage}[t]{0.22\linewidth}
    \centering
    \includegraphics[width=4.cm,height=3.5cm]{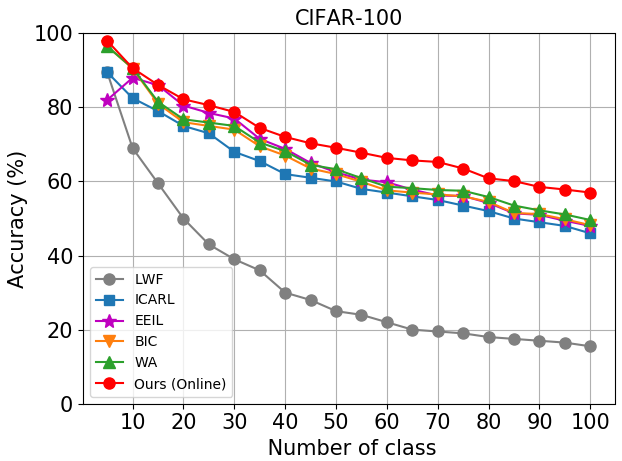}
    \parbox{15.5cm}{\small \hspace{2.cm}(a)}
\end{minipage}
\hspace{2ex}
\begin{minipage}[t]{0.22\linewidth}
    \centering
    \includegraphics[width=4.cm,height=3.5cm]{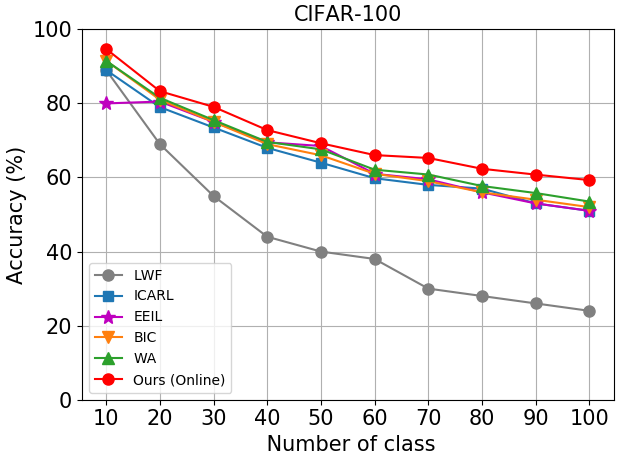}
    \parbox{15.5cm}{\small \hspace{2.cm}(b)}
\end{minipage}
\hspace{2ex}
\begin{minipage}[t]{0.22\linewidth}
    \centering
    \includegraphics[width=4.cm,height=3.5cm]{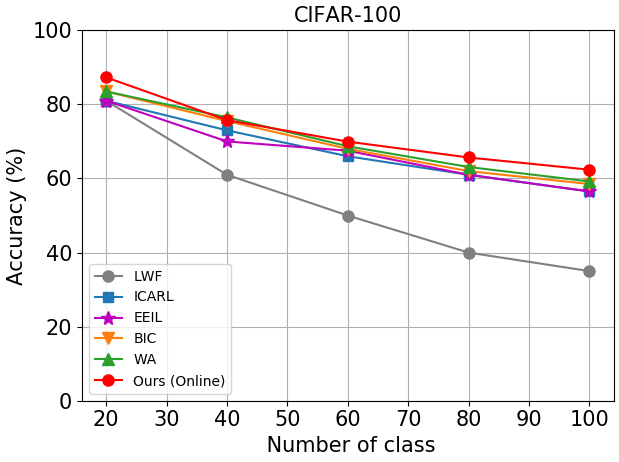}
    \parbox{15.5cm}{\small \hspace{2.cm}(c)}
\end{minipage}
\hspace{2ex}
\begin{minipage}[t]{0.22\linewidth}
    \centering
    \includegraphics[width=4.cm,height=3.5cm]{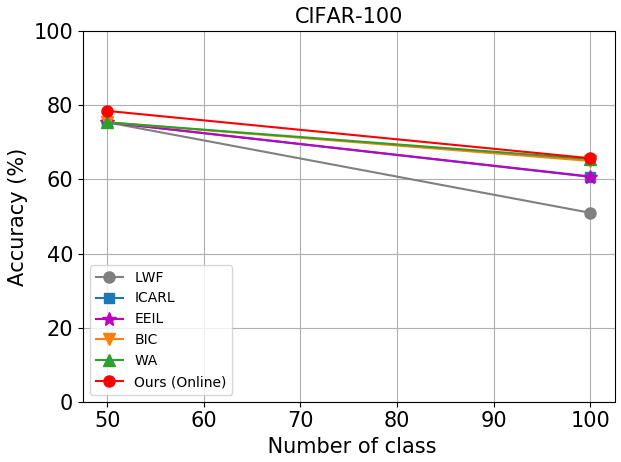}
    \parbox{15.5cm}{\small \hspace{2.cm}(d)}
\end{minipage}
\begin{center}
\vspace{-1.cm}
\caption{\textbf{Results on CIFAR-100} by comparing with offline approaches with step size (a) 5, (b) 10, (c) 20 and (d) 50. Note that only our method is implemented in online. (Best viewed in color)}
\label{fig:offlinecifar}
\end{center}
\end{figure*}

\vspace{-0.2cm}
\section{Experimental Results}
\label{experimental results}
\vspace{-0.2cm}
To show the effectiveness of our proposed approach, we compare with both the state-of-the-art \textit{online methods} following experiment setting similar in~\cite{GEM,A-GEM}, and \textit{offline continual learning methods} as well under benchmark protocol~\cite{ICARL} by varying the incremental step size, which are illustrated in Section~\ref{exp:compare with online} and Section~\ref{exp:compare with offline}, respectively. In Section~\ref{subsec:ablation}, we conduct ablation experiments to validate each component of our propose method. Finally, we study the weight bias problem in online scenario and analyze the storage consumption in Section~\ref{exp:Analysis of Weights Bias, Memory}.
\vspace{-0.1cm}
\subsection{Evaluation Metrics}
\label{exp:metric}
\vspace{-0.1cm}
We focus on continual learning under class-incremental setting as illustrated in Section~\ref{related work}. During inference, the model is evaluated to classify all classes seen so far. We use commonly applied evaluation metrics such as average accuracy (\textit{Avg}) and last step accuracy (\textit{Last}) in this section where \textit{Avg} is calculated by averaging all the accuracy obtained after learning of each task, which shows the overall performance for the entire continual learning procedure. The \textit{Last} accuracy shows the performance after the continual learning for all seen classes. No task index is provided during inference and we ran each experiment five times and report the average Top-1 classification results. 

\subsection{Compare With Online Methods}
\label{exp:compare with online}
\vspace{-0.1cm}
We compare our method with existing \textit{replay-based} online approaches including A-GEM~\cite{A-GEM}, GSS~\cite{NEURIPS2019_e562cd9c_GSS},  MIR~\cite{aljundi_mir}, ASER~\cite{shim2020online_ASER} and GDUMB~\cite{prabhu2020gdumb_online}.

\textbf{Dataset: }We use Split CIFAR-10~\cite{aljundi2019online_splitcifar10}, Split CIFAR-100~\cite{zenke2017continual_splitcifar} and CORE-50~\cite{core50} for evaluation in this part. 
\begin{densitemize}
\itemsep0em 
    \item \textbf{Split CIFAR-10} splits CIFAR-10 dataset~\cite{CIFAR} into 5 tasks with each contains 2 disjoint classes. Each class contains 6,000, $32 \times 32$ RGB images with originally divided 5,000 for training and 1,000 for testing.
    \item \textbf{Split CIFAR-100} contains 20 tasks with non-overlapping classes constructed using CIFAR-100~\cite{CIFAR}. Each task contains 2,500 training images and 500 test images corresponding to 5 classes.
    \item \textbf{CORE-50} is another benchmark dataset for continual learning. For class incremental setting, it is divided into 9 tasks and has a total of 50 classes with 10 classes in the first task and 5 classes in the subsequent 8 tasks. Each class has around 2,400, $128 \times 128$ RGB training images and 900 testing images.
\end{densitemize}

\textbf{Implementation detail: }A small version of ResNet-18 (reduced ResNet-18)~\cite{GEM, A-GEM} pretrained on ImageNet~\cite{IMAGENET1000} is applied as the backbone model for all the compared methods. The ResNet implementation follows the setting as suggested in~\cite{RESNET}. We emphasize that only our method freeze the parameters in backbone network while others do not. We apply SGD optimizer with a mini-batch size of 10 and a fixed learning rate of 0.1. We vary the size of exemplar set for $Q \in \{1000, 2000, 5000, 10000\}$ for comparisons.  
\vspace{-0.2cm}
\subsubsection{Results on Benchmark Datasets}
\label{expres:splitcifar}
\vspace{-0.1cm}
The average accuracy (\textit{Avg}) and last step accuracy \textit{Last} on Split CIFAR-10 and CORE-50 are summarized in Table~\ref{tab:online_cifar_core}. Given different exemplar size $Q$, our method outperforms existing online approaches, especially when $Q$ is smaller by a larger margin, \textit{i.e.}, our method performs better even with limited storage capacity. The reason is that our approach does not solely rely on exemplars to retain old knowledge but maintains the classifier's discriminability for each learned task and makes the prediction through candidates selection and prior incorporation. In addition, our method includes the exemplar augmentation step, which is more effective given limited number of exemplars as analyzed in Section~\ref{subsec:ablation}. In addition, Figure~\ref{fig: online result on cifar} visualizes the results for continual learning of 20 tasks on Split CIFAR-100. The model is evaluated after learning each task on test data belonging to all classes seen far. Our method achieves the best performance for each step and we observe that A-GEM~\cite{A-GEM} does not work well under class-incremental setting, which only use stored exemplars to restrict the update of corresponding parameters while others perform knowledge replay by combining with new class data.

\begin{figure*}[t!]
\vspace{-0.3cm}
\begin{center}
  \includegraphics[width=1.\linewidth]{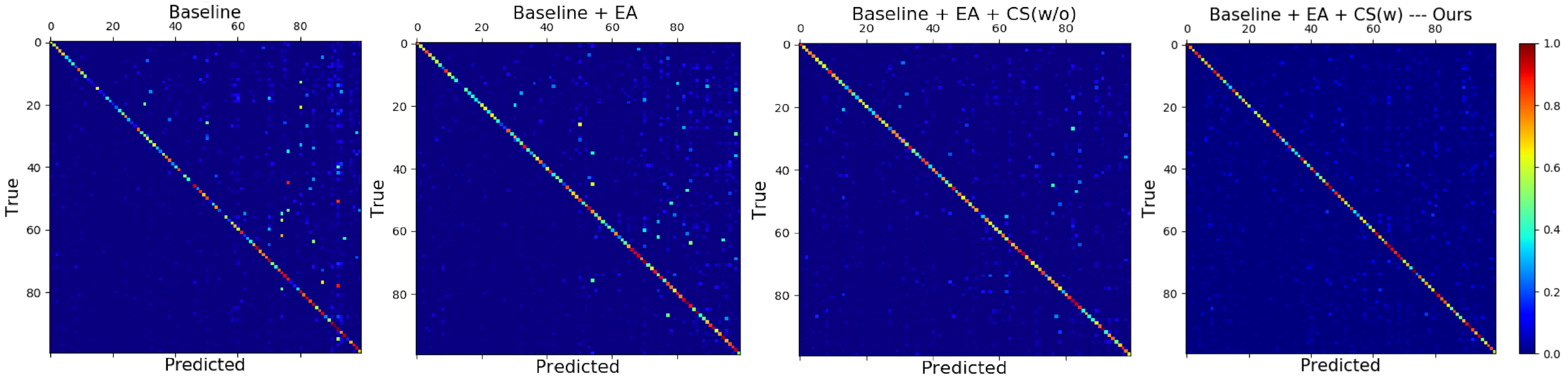}
  \caption{\textbf{Confusion matrices on Split CIFAR-100} for different variants in ablation study. (Best viewed in color)}
  \vspace{-.5cm}
  \label{fig:confusion }
\end{center}
\end{figure*}

\vspace{-0.2cm}
\subsection{Compare With Offline Methods}
\label{exp:compare with offline}
While focusing on online continual learning, we also compare our method with offline continual learning approaches that use each data multiple times to update the model. Although it is widely acknowledged that performance in the online scenario is worse than offline as discussed in~\cite{GEM,prabhu2020gdumb_online} due to the limited number of available new data and each data is observed only once by the model, we show that our method implemented in online scenario is also effective to achieve comparable performance with state-of-the-arts offline approaches including LWF~\cite{LWF}, ICARL~\cite{ICARL}, EEIL~\cite{EEIL}, BIC~\cite{BIC} and WA~\cite{mainatining} following the benchmark protocol similar in~\cite{ICARL}.

\textbf{Datasets: }We use CIFAR-100~\cite{CIFAR} for evaluation and arrange it into splits of 5, 10, 20, and 50 non-overlapped classes, resulting in 20, 10, 5, and 2 tasks, respectively. 

\textbf{Implementation detail: }For experiments on CIFAR-100, we apply ResNet-50~\cite{RESNET} pretrained on ImageNet~\cite{IMAGENET1000} as the backbone model. We apply SGD optimization with mini-batch size of 10 and a fixed learning rate of 0.1 for our method implemented in online scenario. 
For all the experiments, we arrange classes using identical random seed~\cite{ICARL} and use fixed size of exemplar set as $Q= 2,000$. 
\vspace{-0.2cm}

\subsubsection{Results on CIFAR-100}
\label{exp:offlinecifarimagenet}
\vspace{-0.2cm}
We implement our proposed method in online scenario to use each data only once for training (except for the first task, which is learned in offline under this protocol), while all the compared existing methods are implemented in offline for all tasks. The results on CIFAR-100 for each incremental step are shown in Figure~\ref{fig:offlinecifar}. Our method still achieves the best results for all incremental step sizes particularly for smaller step size. One of the reasons is that the weight bias problem becomes more severe with smaller incremental step size (more incremental steps) especially in offline case where the model is updated multiple times for each step, which is analyzed in Section~\ref{exp:Analysis of Weights Bias, Memory}. However, this problem is alleviated in online scenario by our proposed learning strategies to pair each new data with an exemplar as described in Section~\ref{method: learning phase}. Furthermore, our method for inference further mitigate the bias problem by selecting candidates and incorporating prior information using stored exemplars, which is illustrated later in Section~\ref{subsec:ablation}. 


\begin{table}[t!]
    \centering
    \scalebox{.8}{
    \begin{tabular}{c|ccc}
        \hline
        Method & \textbf{CIFAR-10} & \textbf{CIFAR-100} &\textbf{CORE-50}  \\
        \hline
        \hline
        Baseline  & 56.2 & 16.7 & 19.8  \\
        Baseline + EA & 58.9 & 20.1 & 22.4  \\
        Baseline + EA + CS(w/o) & 81.7 & 49.6 & 43.9  \\
        Baseline + EA + CS(w) - Ours & \textbf{84.9} & \textbf{52.0} & \textbf{50.7} \\
        Upper-bound & 92.2 & 70.7 & 67.9 \\
        \hline
    \end{tabular}
    }
    \vspace{-0.2cm}
    \caption{\textbf{Average accuracy (\%) for ablation study on Split CIFAR-10, Split CIFAR-100 and CORE-50}. Best results (except upper-bound) are marked in bold.  }
    \label{tab:ablation}
\end{table}

\begin{table}[t]
    \centering
    \scalebox{.8}{
    \begin{tabular}{c|lll}
        \hline
        Method & \textbf{CIFAR-10} & \textbf{CIFAR-100} &\textbf{CORE-50}  \\
        \hline
        \hline
        Baseline (Q=1,000) & 46.6 & 13.9 & 17.2  \\
        Baseline + EA & 49.8 (\textbf{+3.2}) & 18.5 (\textbf{+4.6}) & 20.6 (\textbf{+3.4})  \\
        \hline
        Baseline (Q=5,000) & 54.9 & 23.8 & 25.4  \\
        Baseline + EA & 56.2 (+1.3) & 25.4 (+1.6)& 26.9 (+1.5) \\
        \hline
        Baseline  (Q=10,000)& 57.2 & 26.8 & 31.4  \\
        Baseline + EA & 58.1 (+0.9)& 27.4 (+0.6)& 31.9 (+0.5) \\
        \hline
    \end{tabular}
    }
    \vspace{-0.2cm}
    \caption{\textbf{Performance of exemplar augmentation step} for the exemplar size $Q \in\{1000, 5000, 10000\}$. Average accuracy (\%) and the corresponding improvements compared with baseline are reported. Highest improvements are marked in bold for each dataset. }
    \label{tab:ablation_expsize}
\end{table}

\subsection{Ablation Study}
\label{subsec:ablation}
\vspace{-0.1cm}
We also conduct ablation study to analyze the effectiveness of each component in our proposed method including \textit{exemplar augmentation in feature space} (EA) and \textit{candidates selection with prior incorporation} (CS) as illustrated in Section~\ref{method: learning phase} and~\ref{method:inference}, respectively. Specifically, we consider the following variants of our method.
\begin{densitemize}
\itemsep0em 
    \item \textbf{Baseline:} remove both CS and EA from our method while keeping exemplar set
    \item \textbf{Baseline + EA:} perform exemplar augmentation
    \item \textbf{Baseline + EA + CS(w/o):} select candidates using stored exemplar but without prior incorporation, which completely trusts the result of PNN by assigning the class of the closest store example as final prediction
    \item \textbf{Baseline + EA + CS(w):} Our proposed method with prior incorporation using Equation~\ref{eq:combine}
\end{densitemize}
We also include \textbf{Upper-bound} for comparison, which is obtained by training a model in non-incremental setting using all training samples from all classes together. We fix the size of exemplar set for $Q = 2,000$ and the average accuracy are summarized in Table~\ref{tab:ablation}. We observe large improvements by adding candidates selection step and our proposed prior incorporation method outperforms directly using PNN output as prediction. The main reason is that the stored feature embeddings extracted by a fixed pre-trained model may not be discriminative enough to make decision especially when there exists obvious distribution difference between the training and testing data as in CORE-50~\cite{core50}, where the data are collected in distinct sessions (such as indoor or outdoor). Therefore, our proposed prior incorporation step mitigate this problem and achieves the best performance. In addition, we also provide confusion matrices as shown in Figure~\ref{fig:confusion } to analyze the results in detail where the \textbf{Baseline} tends to predict new classes more frequently and ours is able to treat new classes and old classes more fairly. Finally, we analyze the exemplar augmentation (EA) by varying exemplar size $Q$ and results are summarized in Table~\ref{tab:ablation_expsize}. Our EA works more efficiently given limited storage capacity, which is one of the most significant constraints to apply continual learning in real world applications. 

\subsection{Weight Bias And Storage Consumption}
\label{exp:Analysis of Weights Bias, Memory}
In this section, we implement additional experiments to show the advantages of our proposed method in online scenario including the analysis of norms of weight vectors in classifier and the comparisons of storage consumption.

\begin{figure}[t]
\begin{minipage}[t]{0.47\linewidth}
    \centering
    \includegraphics[width=4.cm,height=3.5cm]{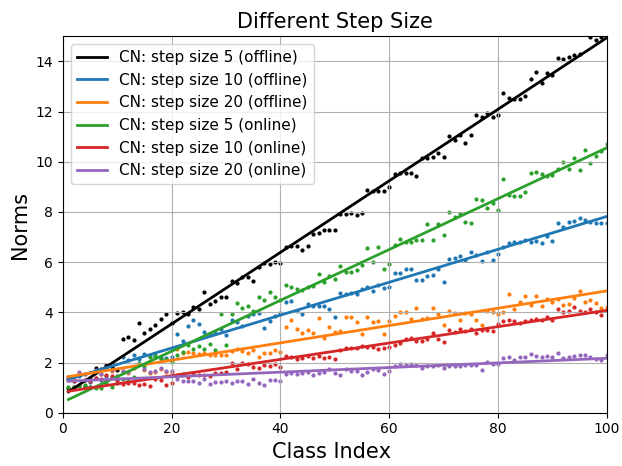}
    \parbox{15.5cm}{\small \hspace{2.cm}(a)}
\end{minipage}
\hspace{2ex}
\begin{minipage}[t]{0.47\linewidth}
    \centering
    \includegraphics[width=4.cm,height=3.5cm]{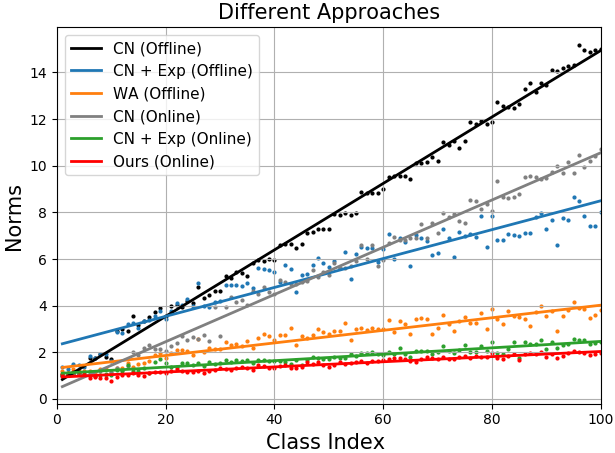}
    \parbox{15.5cm}{\small \hspace{2.cm}(b)}
\end{minipage}
\begin{center}  
   \caption{\textbf{Norms of the weight vectors} for (a) the impact of different step size 5, 10, and 20. (b) Impact of different methods using step size 5. The solid line is obtained by linear least square to show the trend for each case.}
  \vspace{-.9cm}
  \label{fig:weightnorm}
\end{center}
\end{figure}

\textbf{Norms of weight vectors: }One of the main reasons for catastrophic forgetting is the weights in trained model’s FC layer are heavily biased towards new classes, which is already discussed in offline mode~\cite{BIC,mainatining} but lacks sufficient study in online scenario. Therefore, we provide analysis for the impact on biased weights in online and offline scenarios by (1) varying incremental step size and (2) with or without using exemplar set (Exp). For generality, we consider \textbf{CN} and \textbf{CN + Exp} as two baseline methods using regular cross entropy for continual learning without and with exemplars, respectively. We use CIFAR-100 with step size 5, 10 and 20 for experiments. We train 70 epochs in offline as in~\cite{ICARL,EEIL} and 1 epoch in online scenario for each learning step. Results are shown in Figure~\ref{fig:weightnorm}. Each dot corresponds to the norm of the weight vectors in FC layer for each class. For better visualization, we fit the dots using linear least square to show the trend of each method when new classes are added sequentially. 

We observe that the weight bias problem is getting more severe when the number of incremental steps increases, especially in offline case since we repeatedly update model using only new class data. The overall performance in online scenario is much better than offline as each data is used only once for training. 

Next, we show that using exemplars is effective to correct biased weights in both online and offline scenario as indicated by \textbf{CN+EXP} compared to \textbf{CN}. We additionally compare baseline methods with our methods \textbf{Ours} and applying Weight Aligning~\cite{mainatining} denoted as \textbf{WA} for bias correction. The performance of using exemplars in online scenario is even better than applying WA in offline case and our proposed strategy further alleviate this problem. Both analysis explain the larger gains we achieved for smaller step size on CIFAR-100 as discussed in Section~\ref{exp:offlinecifarimagenet}. The comparison between online and offline results also show the potential to address catastrophic forgetting in online scenario with the benefit of reduced weight bias problem.

\textbf{Storage consumption: } Storage requirement poses significant constrains for continual learning in online mode. If we can store all data seen so far without considering storage requirement in real world scenario, then we can easily update the model using all available data. Therefore, we compare the storage consumption of our method with existing approaches to show the significant reduction in storage requirement. Let $S$ denote the image size, $C$ denote the number of total classes seen so far, $Q$ refers to the number of data stored in exemplar set for each class and $D$ denotes the dimension of extracted feature embedding. (1) For methods using original data as exemplars~\cite{prabhu2020gdumb_online,aljundi_mir,NEURIPS2019_e562cd9c_GSS, GEM, A-GEM,chaudhry2019_TINYER,chaudhry2020using_HAL,ICARL,EEIL,BIC,mainatining,ILIO}, the storage requirement for storing data in exemplar set is $O(3\times S^2 \times Q \times C)$. (2) For methods which store statistics of old classes and conduct pseudo rehearsal~\cite{fearnet,cgan}, the total cost is $O(D^2 \times C)$ (3) For our method that store feature embeddings as exemplars, the total storage is $O(D \times C \times Q)$. Therefore, as $Q\ll D < 3\times S^2$, our method requires the least storage while still achieving the best performance.

%% file: 6_conclusion.tex
\vspace{-0.2cm}
\section{Conclusion}
\label{conclusion}
\vspace{-0.2cm}
In summary, we propose a novel and effective method for continual learning in online scenario under class-incremental setting by maintaining the classifier's discriminability for classes within each learned task and make final prediction through candidates selection together with prior incorporation using stored exemplars selected by our online sampler. Feature embedding instead of original data is stored as exemplars, which are both memory-efficient and privacy-preserving for real life applications and we further explore exemplar augmentation in feature space to achieve improved performance especially when given very limited storage capacity. Our method achieves best performance compared with existing online approaches on benchmark datasets including Split CIFAR10, Split CIFAR100 and CORE-50. In addition, we vary the incremental step size and achieves comparable performance even with offline approaches on CIFAR-100. Finally, our analysis on norms of weight vectors in the classifier also shows great potential for addressing catastrophic forgetting in online scenario that can significantly reduce the weight bias problem. Our future work will focus on unsupervised continual learning, which is more realistic and one possible solution is to use pseudo label as recently introduced in~\cite{he2021unsupervised}. 



%% file: 7_supplementary.tex
\section{Estimating $\beta$ Using Pilot Set}
\label{estimate using pilot set}
\begin{figure}[t]
\begin{center}
  \includegraphics[width=1.\linewidth]{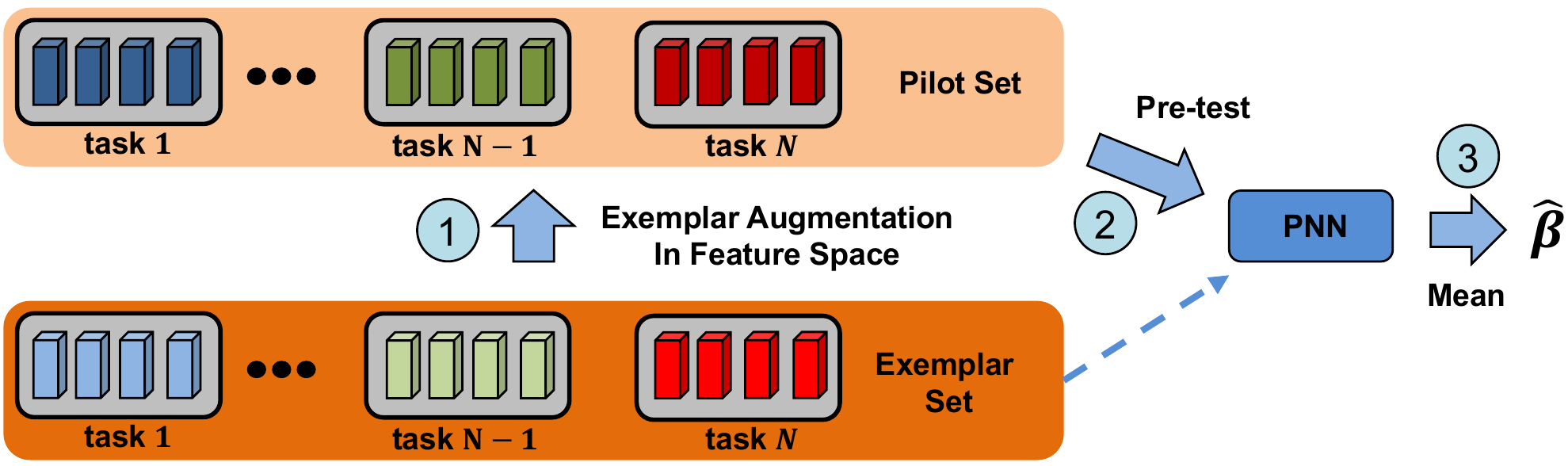}
  \vspace{+0.1cm}
  \caption{\textbf{Estimating $\hat{\beta}$ empirically in three steps.} (1) Constructing a pilot set using all augmented exemplars. (2) Using data in pilot set as input to PNN for pre-test. (3) Calculate the mean value of the difference between maximum and minimum output for each data as the estimated $\hat{\beta}$.}
  \label{fig:pilot set }
\end{center}
\end{figure}

As illustrated in \textbf{Section 3.2}, we combine the extracted and normalized candidates output logits $\hat{S}$ with prior distribution $W$ obtained from PNN using \textbf{Equation 5}, where the dynamic hyper-parameter $\gamma = \frac{\textit{Max}(W)-\textit{Min}(W)}{\beta}$ is used for incorporation. The numerator is calculated by the difference between the maximum and minimum of the prior vector $W$. A larger value of the difference indicates a more confident prior. 
The numerator ranges from $[0, 1]$ and $\beta$ is used for normalization. We use a fixed $\beta = 0.5$ for all experiments shown in the paper. We also provide a simple method that can empirically estimate the $\beta$ before each inference phase. As shown in Figure~\ref{fig:pilot set }, we first construct a pilot set using all augmented exemplars, which are obtained by applying feature space data augmentation as described in \textbf{Section 3.1} on stored exemplars in exemplar set. Then, before each inference phase, we take all the augmented data in the pilot set as input to the PNN classifier for pre-test and we calculate the difference between the maximum and minimum values for each output probability distribution. Finally, we use the mean value of the difference corresponding to all input as the estimated $\hat{\beta}$. 

\begin{figure*}[t]
\centering
\begin{minipage}[t]{0.22\linewidth}
    \centering
    \includegraphics[width=4.cm,height=3.5cm]{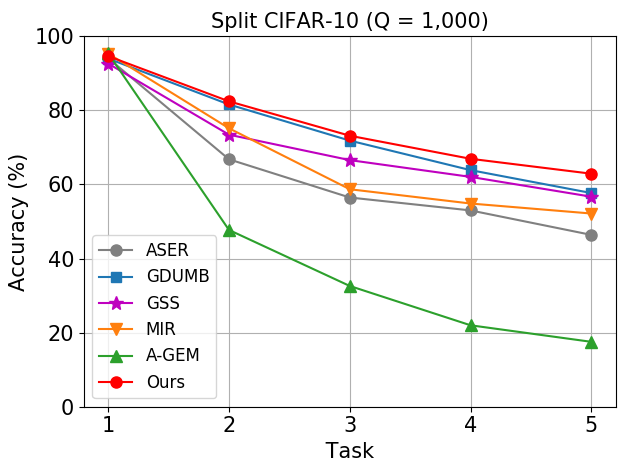}
    \parbox{15.5cm}{\small \hspace{2.cm}(a)}
\end{minipage}
\hspace{2ex}
\begin{minipage}[t]{0.22\linewidth}
    \centering
    \includegraphics[width=4.cm,height=3.5cm]{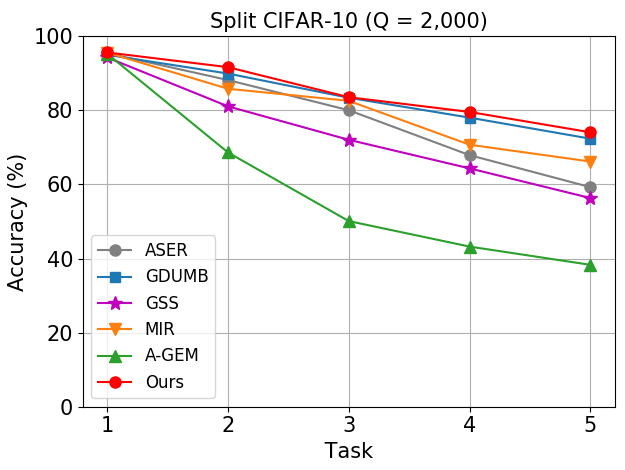}
    \parbox{15.5cm}{\small \hspace{2.cm}(b)}
\end{minipage}
\hspace{2ex}
\begin{minipage}[t]{0.22\linewidth}
    \centering
    \includegraphics[width=4.cm,height=3.5cm]{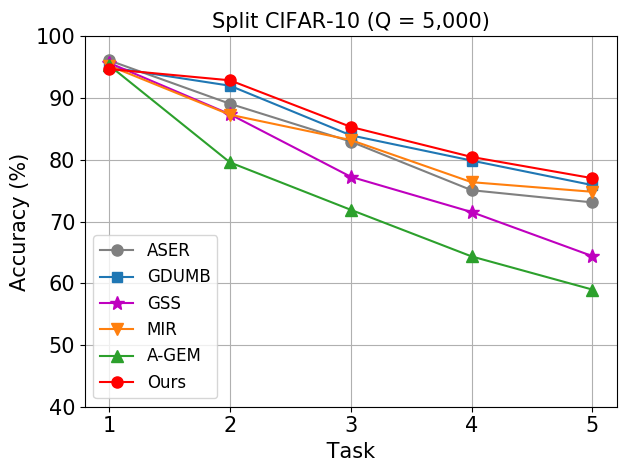}
    \parbox{15.5cm}{\small \hspace{2.cm}(c)}
\end{minipage}
\hspace{2ex}
\begin{minipage}[t]{0.22\linewidth}
    \centering
    \includegraphics[width=4.cm,height=3.5cm]{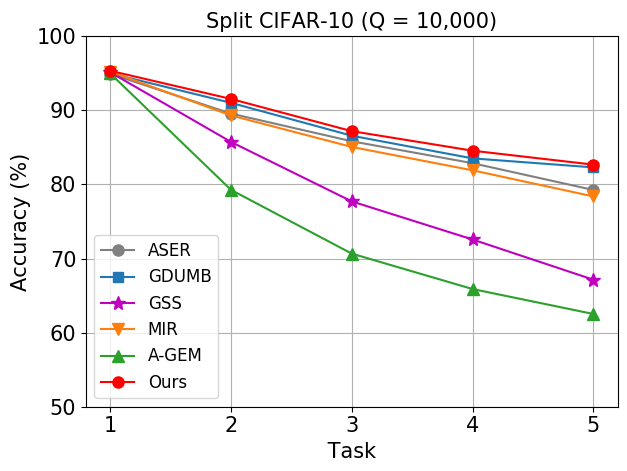}
    \parbox{15.5cm}{\small \hspace{2.cm}(d)}
\end{minipage}
\begin{center}
\vspace{-.8cm}
\caption{\textbf{Results on Split CIFAR-10} by comparing with existing online methods with different exemplar size $Q$. The accuracy is measured after learning of each task on all tasks seen so far. (Best viewed in color)}
\label{fig: cifar10}
\end{center}
\end{figure*}

\begin{figure*}[t]
\centering
\begin{minipage}[t]{0.22\linewidth}
    \centering
    \includegraphics[width=4.cm,height=3.5cm]{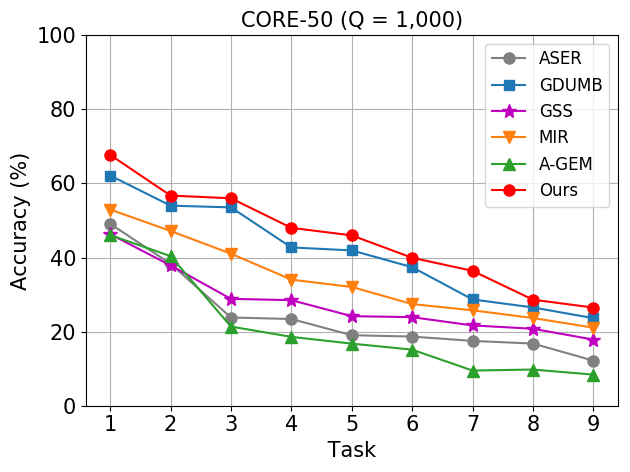}
    \parbox{15.5cm}{\small \hspace{2.cm}(a)}
\end{minipage}
\hspace{2ex}
\begin{minipage}[t]{0.22\linewidth}
    \centering
    \includegraphics[width=4.cm,height=3.5cm]{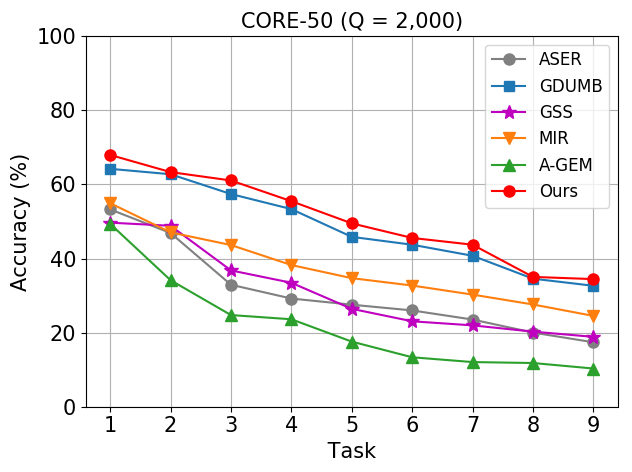}
    \parbox{15.5cm}{\small \hspace{2.cm}(b)}
\end{minipage}
\hspace{2ex}
\begin{minipage}[t]{0.22\linewidth}
    \centering
    \includegraphics[width=4.cm,height=3.5cm]{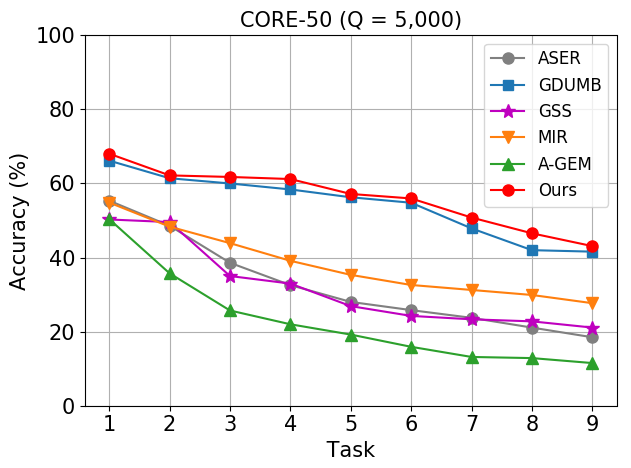}
    \parbox{15.5cm}{\small \hspace{2.cm}(c)}
\end{minipage}
\hspace{2ex}
\begin{minipage}[t]{0.22\linewidth}
    \centering
    \includegraphics[width=4.cm,height=3.5cm]{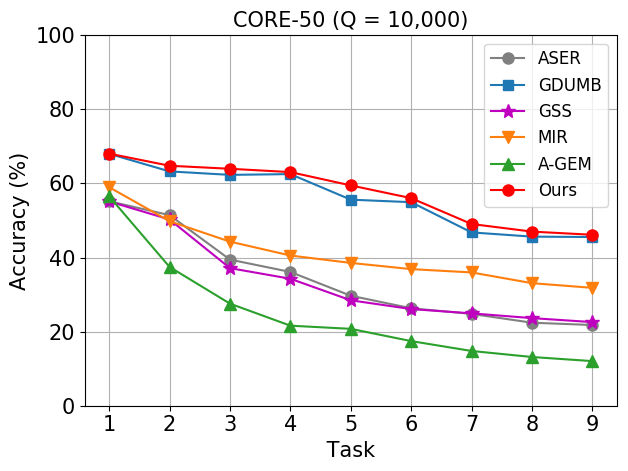}
    \parbox{15.5cm}{\small \hspace{2.cm}(d)}
\end{minipage}
\begin{center}
\vspace{-.8cm}
\caption{\textbf{Results on CORE-50} by comparing with existing online methods with different exemplar size $Q$. The accuracy is measured after learning of each task on all tasks seen so far. (Best viewed in color)}
\label{fig: core50}
\end{center}
\end{figure*}

\begin{table*}[t!]
    \vspace{-0.3cm}
    \centering
    \scalebox{1.}{
    \begin{tabular}{ccccccccc}
        \hline
        Datasets & \multicolumn{8}{c}{\textbf{Split CIFAR-100}}  \\
        \hline
        Size of exemplar set & \multicolumn{2}{c}{$Q = 1,000$} & \multicolumn{2}{c}{$Q = 2,000$} & \multicolumn{2}{c}{$Q = 5,000$} & \multicolumn{2}{c}{$Q = 10,000$}\\
        \hline
        Accuracy(\%) & Avg & Last & Avg & Last & Avg & Last & Avg & Last \\
        \hline
        \hline
        A-GEM~\cite{A-GEM}  & 13.9 & 4.3 & 14.1& 4.83& 13.8 & 4.4 & 13.9 & 4.5 \\
        MIR~\cite{aljundi_mir} & 37.4 & 19.1& 41.9 & 26.1 & 44.4 & 27.9 & 46.3 & 34.0 \\
        GSS~\cite{NEURIPS2019_e562cd9c_GSS} & 24.2 & 12.9 & 26.9 & 16.7 & 31.5 & 23.1 & 43.2 & 23.3 \\
        ASER~\cite{shim2020online_ASER} & 29.9 & 21.6 & 40.9 & 27.8 & 41.2 &29.5 & 43.3 & 30.8\\
        GDUMB~\cite{prabhu2020gdumb_online} & 38.3 & 20.7 & 39.2 & 26.4 & 43.8 & 31.6 & 52.8 & 39.2\\
        \hline
        Ours & \textbf{50.4} & \textbf{35.6} & \textbf{51.8} & \textbf{36.5} & \textbf{54.1} & \textbf{37.4} & \textbf{54.9} & \textbf{40.2} \\
        \hline
    \end{tabular}
    }
    \vspace{-0.2cm}
    \caption{\textbf{Average accuracy and Last step accuracy on Split CIFAR-100}. Best results marked in bold.  }
    \label{tab:cifar100}
\end{table*}

\section{Implementation Detail and Additional Experimental Results}
\label{sec: detail and results}
\subsection{Comparison With Online Methods}
\label{subsec: compare with online}

In \textbf{Section 4.2}, we compare our online method with existing online work including A-GEM~\cite{A-GEM}, MIR~\cite{aljundi_mir}, GSS~\cite{NEURIPS2019_e562cd9c_GSS}, ASER~\cite{shim2020online_ASER} and GDUMB~\cite{prabhu2020gdumb_online}. In this part, we show detail settings for all compared methods with additional detail experimental results on Split CIFAR-10, Split CIFAR-100 and CORE-50. 
\begin{itemize}
    \item A-GEM is an improved version of GEM~\cite{GEM}, which address catastrophic forgetting by restricting the update of parameters in the model that are important for learned tasks. It ensures that the average loss calculated by using stored old task exemplars does not increase for each training step 
    \item MIR refers to Maximally Interfered Retrieval, which applies reservoir sampling to select exemplars for knowledge replay. During training phase, it selects stored samples that are maximally interfered with the largest increase of loss through virtual parameter update criterion for the incoming new data  
    \item GSS aims to store exemplars that have diversity of gradient directions. It calculates the score for each exemplar through maximal cosine similarity in gradient space between that exemplar and a randomly constructed exemplar subset where the sample of lower score will be kept in exemplar set
    \item ASER is a recently proposed online method, which provides a novel scoring method, \textit{Adversarial Shapley Value}, to select exemplars that can better maintain the decision boundaries for all classes learned so far while encouraging plasticity and optimal learning of current new class decision boundaries
    \item GDUMB is another the most recent online approach, which applies a balanced greedy sampler to store as much as learned data it allowed and trains a classifier during inference using stored data only 
\end{itemize}

\textbf{Additional results: }in \textbf{Section 4.2.1} we show results on benchmark datasets by comparing with above online approaches. In this part we provide (1) the visualization of performance for evaluated aftering of each task on Split CIFAR-10 and CORE-50, which are shown in Figure~\ref{fig: cifar10} and Figure~\ref{fig: core50}, respectively. (2) The average accuracy and last step accuracy for Split CIFAR-100 is summarized and shown in Table~\ref{tab:cifar100}.

\begin{table*}[t!]
    \centering
    \scalebox{1.}{
    \begin{tabular}{ccccccccc}
        \hline
        Datasets & \multicolumn{8}{c}{\textbf{CIFAR-100}}\\
        \hline
        Step size& \multicolumn{2}{c}{5} & \multicolumn{2}{c}{10} & \multicolumn{2}{c}{20} & \multicolumn{2}{c}{50} \\
        \hline
        Accuracy(\%) & Avg & Last & Avg & Last & Avg & Last & Avg & Last \\
        \hline
        \hline
        LWF~\cite{LWF}  & 29.7 & 15.5 & 39.7 & 24.0 & 47.1 &35.1 & 52.6 & 52.6\\
        ICARL~\cite{ICARL} & 59.7 & 46.0 & 61.6& 51.0 & 63.3 & 56.5 & 60.7 & 60.7 \\
        EEIL~\cite{EEIL} & 63.4 & 48.0 & 63.6& 51.5 & 64.2 & 57.1 & 61.2 & 61.2 \\
        BIC~\cite{BIC} & 62.1 & 48.2& 63.5&52.0 & 65.1 &58.5  & 64.9 & 64.9 \\
        WA~\cite{mainatining} & 62.6 & 49.6& 64.5& 53.5& 66.6 & 59.2 & 65.1 & 65.1 \\
        \hline
        Ours  & \textbf{70.6} & \textbf{57.0} & \textbf{69.9}&\textbf{59.3} & \textbf{70.5} & \textbf{62.4 }&\textbf{65.7} & \textbf{65.7} \\
        \hline
    \end{tabular}
    }
    \vspace{-0.2cm}
    \caption{\textbf{Average accuracy and Last step accuracy on CIFAR-100} by comparing with offline approaches. Note that the compared methods are implemented in offline while ours is implemented in online. Best results marked in bold.}
    \label{tab:offline_all3}
\end{table*}

\subsection{Comparison With Offline Methods}
\label{subsec: compare with offline}
In \textbf{Section 4.3}, we compare our method implemented in online scenario with existing methods implemented in offline scenario. In this part, we show detail settings for all compared offline methods including LWF~\cite{LWF}, ICARL~\cite{ICARL}, EEIL~\cite{EEIL}, BIC~\cite{BIC} and WA~\cite{mainatining}. 
\begin{itemize}
    \item LWF proposes to use knowledge distillation loss~\cite{KD} using a fixed teacher model from last learning step to mitigate forgetting. Note that it is originally designed for task-incremental problem, so for all experiments shown in paper, we use a variant of it introduced in~\cite{ICARL}.
    \item ICARL also adopts distillation loss but it additionally selects fixed number of learned data as exemplars for knowledge replay through Herding algorithm~\cite{HERDING}. A Nearest Class Mean (NCM) classifier is used to make final prediction during inference phase  
    \item EEIL improves upon ICARL by implementing in an end-to-end fashion using softmax classifier and it also applies data augmentation and balanced fine-tuning
    \item BIC is the first one to address catastrophic forgetting through bias correction in classifier, which applies an additional linear bias correction layer to estimate parameters by constructing a balanced validation set
    \item WA is the most recent work that targets on bias correction, which calculates the norms of weight vectors corresponding to old and new classes and use the ratio of mean value to correct biased output logits
    does not require additional parameters 
\end{itemize}

\textbf{Implementation detail: }We apply SGD optimizer with mini-batch size of 10 and a fixed learning rate of 0.1 for our method implemented in online scenario for CIFAR-100. Each data is used only once for training, \textit{i.e. }training epoch is 1. The implementation of all existing methods follows their own repositories and we summarize the training epoch and batch size as shown in Table~\ref{tab:batch_epoch}. Our method requires the least number of available data (batch size: 10) and use each data only once to update (epoch: 1) while achieving promising results as illustrated in \textbf{Section 4.3}. 

\textbf{Additional results: }in \textbf{Section 4.3.1} we visualize the result evaluated after each incremental step on CIFAR-100 with various step sizes by comparing with above offline approaches. In this part we provide the average accuracy and last step accuracy as summarized in Table~\ref{tab:cifar100}. We want to emphasize that the focus of our paper is to introduce online continual learning method, but we surprisingly found that our performance is even better than offline approaches since it is widely acknowledged that performance in the online scenario is worse than offline when using the same method as discussed in~\cite{GEM,prabhu2020gdumb_online} due to the limited number of available new data and each data is observed only once by the model. Therefore, we also presents the results on CIFAR-100 by comparing with offline approaches and then investigate why this happens by analyzing the weight bias problem as shown in \textbf{Section 4.5} and Section~\ref{bias analysis}. 

\begin{table}[t!]

    \centering
    \scalebox{1.}{
    \begin{tabular}{c|cc|cc}
        \hline
        Datasets & \multicolumn{2}{c}{\textbf{CIFAR-100}} & \multicolumn{2}{|c}{\textbf{ImageNet}}\\
        \hline
         Training & Batch Size & Epoch & Batch Size & Epoch \\
        \hline
        \hline
        LWF~\cite{LWF}  & 128 & 70 & 128 & 60 \\
        ICARL~\cite{ICARL} & 128 & 70 & 128& 60 \\
        EEIL~\cite{EEIL} & 128 & 70 & 128 & 70  \\
        BIC~\cite{BIC} & 128 & 250 & 256 & 100 \\
        WA~\cite{mainatining} & 32 & 250 & 256 & 100 \\
        \hline
        Ours  & 10 & 1 & 10 & 1\\
        \hline
    \end{tabular}
    }
    \vspace{+0.2cm}
    \caption{\textbf{Training batch size and epochs} for all compared existing offline approaches and our online method.}
    \label{tab:batch_epoch}
\end{table}

\begin{figure}[t]
\begin{minipage}[t]{0.47\linewidth}
    \centering
    \includegraphics[width=4.cm,height=3.5cm]{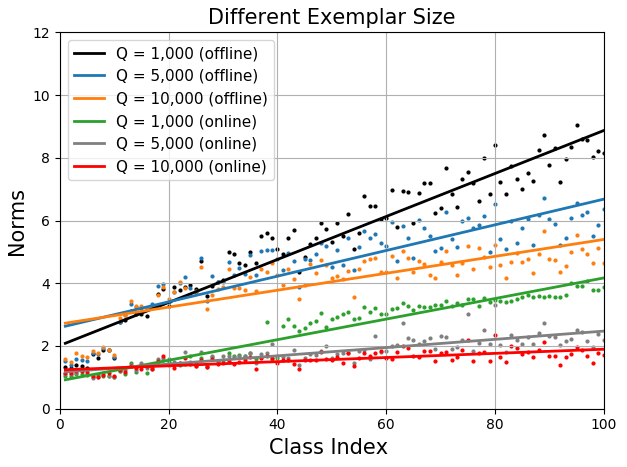}
    \parbox{15.5cm}{\small \hspace{2.cm}(a)}
\end{minipage}
\hspace{2ex}
\begin{minipage}[t]{0.47\linewidth}
    \centering
    \includegraphics[width=4.cm,height=3.5cm]{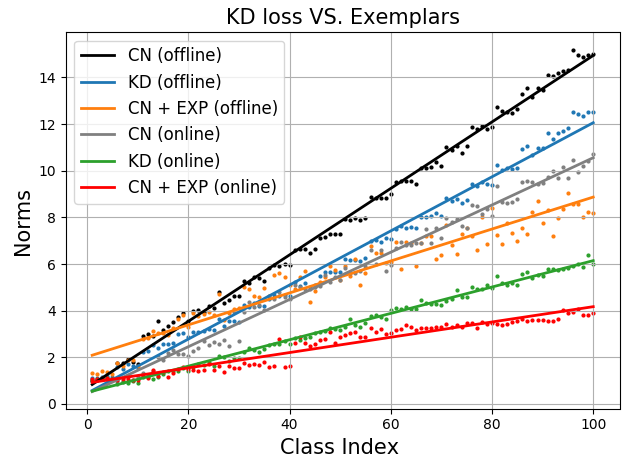}
    \parbox{15.5cm}{\small \hspace{2.cm}(b)}
\end{minipage}
\begin{center}  
  \caption{\textbf{Norms of the weight vectors on CIFAR-100} with step size 5 for (a) the impact of different exemplar size $Q$. (b) Impact of using knowledge distillation loss. The solid line is obtained by linear least square to show the trend for each case.}
  \label{fig:norm}
\end{center}
\end{figure}

\section{Bias Weight Analysis}
\label{bias analysis}
In \textbf{Section 4.5}, we provide the comparison results between online and offline for weight bias problem by varying the step size and using exemplars. In this part, we show additional results for (1) varying the size of exemplar set and (2) using knowledge distillation loss~\cite{KD} for bias correction. For the first part, we apply the baseline method using cross-entropy loss to update model $\textbf{(CN)}$ for experiments. For the second part, we additionally use exemplars denoted as \textbf{CN + EXP}, and \textbf{KD} refers to replacing cross-entropy with knowledge distillation loss.

\textbf{The influence of exemplar size: }We vary the exemplar size $Q \in \{1,000, 5,000, 10,000\}$. Figure~\ref{fig:norm}(a) shows the results on CIFAR-100 with step size 5. As expected, we observe that the biased weights are alleviated when increasing the number of exemplars in both online and offline scenarios. In addition, the overall performance in online scenario is much better than in offline and using $Q=10,000$ in online mode almost resolves this problem. However, the storage capacity is also a significant constraint for continual learning especially in online scenario, so there is a trade off between storage consumption and the performance. As shown in \textbf{Section 4.2} and \textbf{Section 4.3}, our method use the least storage while achieving the best performance.

\textbf{The influence of knowledge distillation loss: }We compare the effectiveness of using exemplars with using knowledge distillation loss for bias correction in both online and offline scenarios. We set $Q=1,000$ for baseline method using exemplars and the results on CIFAR-100 with step size 5 is shown in Figure~\ref{fig:norm}(b). Although only small number of exemplars are used ($Q=1,000$), the performance of \textbf{CN + EXP} is better than using knowledge distillation loss (\textbf{KD}) in online and offline scenarios. In addition, both exemplars and distillation loss become more efficient in online case for bias correction, showing great potential to address catastrophic forgetting in online scenario.